\documentclass{ecai} 
\usepackage{latexsym}
\usepackage{amssymb}
\usepackage{amsmath}
\usepackage{amsthm}
\usepackage{booktabs}
\usepackage{enumitem}
\usepackage{graphicx}
\usepackage{color}
\usepackage{makecell}
\usepackage{multirow}
\usepackage{siunitx}
\newcommand{\BibTeX}{B\kern-.05em{\sc i\kern-.025em b}\kern-.08em\TeX}

\begin{document}

%%%%%%%%%%%%%%%%%%%%%%%%%%%%%%%%%%%%%%%%%%%%%%%%%%%%%%%%%%%%%%%%%%%%%%%%

\begin{frontmatter}

%%% Use this command to specify your submission number.
%%% In doubleblind mode, it will be printed on the first page.

\paperid{7846} 

%%% Use this command to specify the title of your paper.

\title{The Influence of Human-inspired Agentic Sophistication in LLM-driven Strategic Reasoners}

%%% Use this combinations of commands to specify all authors of your 
%%% paper. Use \fnms{} and \snm{} to indicate everyone's first names 
%%% and surname. This will help the publisher with indexing the 
%%% proceedings. Please use a reasonable approximation in case your 
%%% name does not neatly split into "first names" and "surname".
%%% Specifying your ORCID digital identifier is optional. 
%%% Use the \thanks{} command to indicate one or more corresponding 
%%% authors and their email address(es). If so desired, you can specify
%%% author contributions using the \footnote{} command.

\author[A]{\fnms{Vince}~\snm{Trencsenyi}\orcid{0009-0009-4560-7571}\thanks{Corresponding Author. Email: vince.trencsenyi@rhul.ac.uk}}
\author[A]{\fnms{Agnieszka}~\snm{Mensfelt}\orcid{0000-0002-2385-2017}}
\author[A]{\fnms{Kostas}~\snm{Stathis}\orcid{0000-0002-9946-4037}} 

\address[A]{Department of Computer Science, Royal Holloway University of London}

%%% Use this environment to include an abstract of your paper.

\begin{abstract}
The rapid rise of large language models (LLMs) has shifted artificial intelligence (AI) research toward agentic systems, motivating the use of weaker and more flexible notions of agency. However, this shift raises key questions about the extent to which LLM-based agents replicate human strategic reasoning, particularly in game-theoretic settings. In this context, we examine the role of {\em agentic sophistication} in shaping artificial reasoners' performance by evaluating three agent designs: a simple game-theoretic model, an unstructured LLM-as-agent model, and an LLM integrated into a traditional agentic framework. Using guessing games as a testbed, we benchmarked these agents against human participants across general reasoning patterns and individual role-based objectives. Furthermore, we introduced obfuscated game scenarios to assess agents' ability to generalise beyond training distributions. Our analysis, covering over 2000 reasoning samples across 25 agent configurations, shows that human-inspired cognitive structures can enhance LLM agents' alignment with human strategic behaviour. Still, the relationship between agentic design complexity and human-likeness is non-linear, highlighting a critical dependence on underlying LLM capabilities and suggesting limits to simple architectural augmentation.
\end{abstract}

\end{frontmatter}

%%%%%%%%%%%%%%%%%%%%%%%%%%%%%%%%%%%%%%%%%%%%%%%%%%%%%%%%%%%%%%%%%%%%%%%%

\section{Introduction}

The transition from LLMs to agentic AI prompts a fundamental reconsideration of traditional models of agency and multi-agent systems~\cite{huang2024understanding}. Earlier conceptions of agency~\cite{Wooldridge_Jennings_1995} typically invoked the intentional stance, framing agents as belief holding entities that are capable of deliberation and reasoning with them to generate and enact their intentions~\cite{bdi-0,bdi-kr}. In contrast, the growing success of LLMs has led to their increasing characterisation as inherently agentic~\cite{liu2023agentbenchevaluatingllmsagents} -- evaluated not only as capable of reasoning, but also as demonstrating cognitive abilities~\cite{amirizaniani2024llms,strachan2024testing,ullman2023large}, such as forming recursive (or nested) beliefs about others' mental states, as if possessing a theory of mind~\cite{WIMMER1983theoryofmind}, much like humans~\cite{sharma2023investigating,dasgupta2022language}.

As part of a broader research agenda on using agentic AI for game-theoretic social simulations~\cite{Shults2025}, we are interested in developing agentic AIs exhibiting human-like strategic reasoning for navigating scenarios of varying complexity, anticipating others' actions, and selecting responses accordingly~\cite{Rasmusen2006introtogametheory}. Experimental game theory has yielded extensive evidence documenting human performance in such recursive reasoning tasks~\cite{costagomes2006guessinggames, duffy1997robustness,duffy2016macroeconomics,nagel08, nagel95,NAGEL2008391}. Building on this foundation, studies have begun to examine the capabilities of LLMs in competitive environments~\cite{mensfelt2024logicenhancedlanguagemodelagents,mensfelt2024autoformalizing,mao2024alympicsllmagentsmeet,duan2024gtbench,zhang2024klevelreasoningestablishinghigher,gandhi2023strategicreasoninglanguagemodels}. Although agentic applications are proliferating rapidly~\cite{wang2024survey} -- from simple question answering to highly structured agentic workflows -- a significant gap remains in understanding \textit{agentic sophistication}, viz., how to rank the many ways LLM capabilities can be employed in constructing agents, and the way such sophistication shapes LLM-based agents to replicate the variety of human strategic reasoning.

Not only is agentic sophistication essential  but so too is the challenge of integrating diverse agentic AI systems within unified simulation environments. While several frameworks evaluate LLM-based agents through game-based interactions~\cite{wu2024smartplaybenchmarkllmsintelligent,liu2023agentbenchevaluatingllmsagents} and specialised game-theoretic benchmarks~\cite{duan2024gtbench,mensfelt2024logicenhancedlanguagemodelagents,mao2024alympicsllmagentsmeet,hua2024gametheoreticllmagentworkflow}, no formally defined and widely adopted multi-agent simulation framework yet exists that can host heterogeneous agent architectures within a standardised environment. Instead, a variety of divergent benchmarking approaches have emerged, making it difficult to objectively determine which agents are most suitable for particular applications~\cite{kapoor2024aiagentsmatter}.

In addition to integration challenges, the black-box nature of LLMs introduces further concerns for agentic applications, particularly regarding reproducibility, explainability, and validation of agent behaviour~\cite{li2024survey}. These issues are especially acute in agent-based models designed for human-centered domains, where trustworthiness and safety are critical~\cite{larooij2025largelanguagemodelssolve,anwar2024foundationalchallengesassuringalignment}. Although several approaches have been proposed to address validation -- such as formal representations, out-of-distribution analysis, and human-in-the-loop evaluation~\cite{collins2022structuredflexiblerobustbenchmarking,mensfelt2024autoformalizing,mensfelt2024autoformalizationgamedescriptionsusing,shankar2024validates} -- many LLM-driven applications continue to lack systematic and rigorous validation mechanisms.

While these challenges remain underexplored and unresolved in game-theoretic social simulations, our work directly addresses them through the following contributions. We introduce guessing games as a testbed for strategic reasoning and present a human dataset, segmented by sub-populations, as a benchmark. Building on a formally defined multi-agent framework grounded in traditional agent concepts, we implement diverse agentic models and use this system to run agent simulations of two-player guessing games. Through these experiments, we explore the relationship between human-inspired sophistication and artificial reasoners, targeting two core problems: (1) replicating general patterns of human reasoning and (2) approximating individual-level variation. Our analysis spans over 2000 reasoning samples from 25 agent configurations, enabling detailed comparison of human and agent performance across multiple dimensions. Additionally, we introduce an obfuscated game scenario and an out-of-sample validation treatment to address potential biases from LLMs’ exposure to game-theoretic materials during training. Finally, we summarise our key findings and outline promising directions for future research.

\section{Human Reasoning in 2-Player Guessing Games}
Game theory provides a formal foundation for modelling strategic interactions among rational agents~\cite{Osborne2004}. \textit{Guessing games}, or \textit{p-Beauty Contests}~\cite{keynes1937general}, are a canonical paradigm for evaluating players' strategic reasoning, particularly their ability to anticipate others' decisions~\cite{nagel95}. In these games, $n$ players assumed strangers, simultaneously select a number from a defined range. The winner is the player whose choice is closest to $p$ times the mean of all selections. 

A solution in guessing games is typically modelled using \textit{k-level reasoning theory}~\cite{camerer2004cognitive}, which assumes that each level-$k$ player believes others reason at most at level $k-1$ and best responds accordingly. This recursive process is formalised as: $a_k = p^k \cdot a_0$, where $a_0$ denotes the choice of 0-level players and $a_k$ the response of level-$k$ players.
While in $n$-player repeated games players often iteratively reject weakly dominant strategies, gradually driving the average guess toward zero, the dynamics change considerably in two-player settings. Specifically, when $p < 1$, choosing $0$ becomes a strictly dominant strategy, as the lower guess will always win. Despite this theoretical simplicity, empirical studies show that players frequently select dominated strategies~\cite{nagel95}. 

We focus on \textit{two-player one-shot guessing games}, which offer several advantages. First, they remove the complexity associated with iterative learning and reasoning about future actions, thereby simplifying the analysis. Second, they likely reduce biases in large language models introduced by exposure to $n$-player beauty contest scenarios during pretraining. We adopt the two-player one-shot guessing game dataset from~\cite{nagel95}, which includes responses from two distinct participant cohorts:
\begin{itemize}
    \item \textbf{Students}: First-year undergraduates, assumed to have no prior exposure to game theory or strategic reasoning.
    \item \textbf{Experts}: Professionals recruited at economics and psychology decision-making conferences, assumed to have relevant knowledge of game-theoretic concepts.
\end{itemize}

\begin{figure}[ht]
\centering
\includegraphics[width=\linewidth]{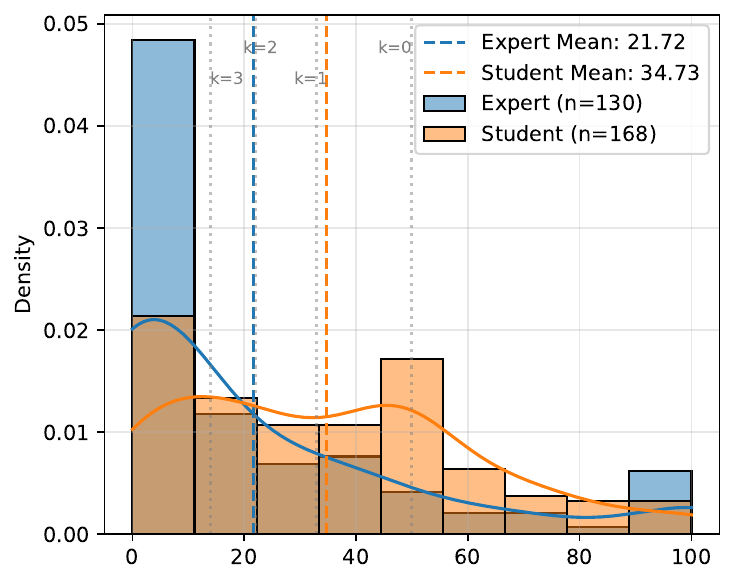}
\caption{Distribution of human guesses per student and expert cohort. Vertical grey lines mark the standard k-levels for guessing games.}
\label{fig:humans}
\end{figure}

Theoretical expectations suggest that the expert group is more likely to identify the dominant strategy than students. This is reflected in our dataset, as $11.9\%$ of students and $37.7\%$ of experts chose $0$. Furthermore, Figure~\ref{fig:humans} supports the expected intergroup performance difference: the expert cohort displays a substantially lower mean guess and greater density over lower values. Experts also exhibit higher skewness ($1.50$ vs. $0.55$). A Levene's test found no significant difference in variances between the groups ($p = 0.4351$), justifying the use of a standard independent samples $t$-test. The test confirms that the difference in means is statistically significant ($t = 4.04$, $p = 0.0001$).

\section{Methods}

\subsection{Multi-agent Simulation Framework}

We build on our earlier work~\cite{trencsenyi2025approximating}, which introduced a centralised multi-agent framework combining the Observe–Orient–Decide–Act (OODA) control loop~\cite{OODA_agents} with multi-level hypergames~\cite{Wang1988} to support recursive reasoning. For the present study, we define a streamlined variant that omits explicit recursive evaluation and instead emphasises the core interaction cycle between agents.

\paragraph{Framework Overview}
The framework centres on an \emph{umpire}~$\upsilon$, conceptualised as a pseudo-player~\cite{Rasmusen2006introtogametheory}. The umpire initialises each player with a context $c_i \in C$ and an instruction model $m_i \in M$, where $C$ is the set of agent contexts and $M$ the set of instruction templates. The umpire also manages gameplay, pairing agents and orchestrating their participation in the guessing game.

\paragraph{Agent Processes}
Given a natural language description $x \in X$, where $X$ is the space of game descriptions, each player $i$ applies an \emph{interpretation function}:
\[
I : X \times C \times M \rightarrow \mathbb{Z},
\]
which encapsulates the orient and decide phases of the agent cycle. We refer to $I$ as the player’s \emph{cognitive process}, mapping $(x,c_i,m_i)$ to a numerical guess. Different implementations of $I$ capture distinct forms of reasoning, and their comparison constitutes the focus of our study on \emph{agentic sophistication}.

\paragraph{Game Formalisation}
Following~\cite{trencsenyi2025approximating}, we adopt a perspective-based formalisation of two-player guessing games. A game is defined as
\[
G = (N, p, A, U),
\]
where $N=\{i,j\}$ are the two players, $p=\tfrac{2}{3}$ is the target multiplier, and $A = A_i = A_{-i} \subseteq \mathbb{Z}$ is a shared action space, typically $A=\{0,1,\dots,100\}$. The utility function $U:A \times A \to \mathbb{R}$ is
\begin{equation}
U_i(a_i, a_{-i}) =
\begin{cases}
1 & \text{if } |a_i - p\mu| < |a_{-i} - p\mu|, \\
0.5 & \text{if } |a_i - p\mu| = |a_{-i} - p\mu|, \\
0 & \text{otherwise},
\end{cases}
\end{equation}
where $\mu = \tfrac{a_i+a_{-i}}{2}$ is the mean guess.

\paragraph{Roadmap}
In the following sections, we instantiate this framework with several agent types. We begin with a game-theoretic benchmark (EWA), then introduce two LLM-based artificial reasoners, each implementing $I$ with different levels of conceptual sophistication.

\subsection{Experience Weighted Attraction}

We choose as a benchmark the self-tuning experience weighted attraction model (EWA) as a single-parameter game-theoretic learning algorithm designed to produce results that mimic human behaviour~\cite{ho2007self}. The model relies on three specific modeling features that assign attraction values to the agent's strategies, which are (i) reinforced by earned payoffs, (ii) simulated by considering foregone payoffs, and (iii) decayed by a memory factor. The model is driven by the update function $\mathcal{A}_j^i(t)$ assigning attention values at each step $t$ to a player $i$' strategy $j$ -- denoted by $s_i^j$ --, defined as follows:

\begin{equation}\label{eq:main}
    A^j_i(t) = \frac{\text{Memory}^j_i(t) + \text{Reinforce}^j_i(t)}{N(t-1) \cdot \phi(t) \cdot (1-\kappa)}.
\end{equation}
The \textit{Memory} component integrates a cumulative experience weight $N$~\cite{CAMERER2002137} and defines how past experience is carried forward:

\begin{equation}\label{eq:memory}
    \text{Memory}^j_i(t) = \phi(t) \cdot N(t-1) \cdot \mathcal{A}^j_i(t-1),
\end{equation}
where $\phi$ denotes a change detector function
\begin{equation}
    \phi(t)=1-\frac{1}{2}S_i(t),
\end{equation}
specifying how much the agent's current observation $h$ differs from the accumulated past experience $r$, encapsulated in the \textit{Surprise index} $S_i$:
\begin{equation}\label{fig:supr}
    S_i(t)=\sum^{m_{-i}}_{k=1}{(h^k_i(t)-r^k_i(t))^2}.
\end{equation}
Given player payoff $\pi_i$, EWA's choice reinforcement is modeled as:
\begin{equation}
    \text{Reinforce}^j_i(t) = [\delta+(1-\delta) \cdot \mathcal{I}(s^j_i,s_i(t))] \cdot \pi_i(s^j_i,s_{-i}(t)),
\end{equation}
where $\delta$ denotes the simulation of ``would have earned'' payoffs:
\begin{equation}\label{eq:foregone}
    \delta_{ij}(t)= \begin{cases}
        1 & \text{if } \pi_i(s^j_i,s_{-i}(t)) \geq \pi_i(t),\\
        0 & \text{otherwise.}
    \end{cases}
\end{equation}
Finally, A softmax exponential form prediction function is used over attractions to sample next period play, leaving $\lambda$ -- controlling sensitivity -- as a single free parameter to be tuned:

\begin{equation}
    P^j_i(t+1) = \frac{e^{\lambda \cdot \mathcal{A}^j_i(t)}}{\sum^{m_i}_{k=1}{e^{\lambda \cdot \mathcal{A}^k_i(t)}}}.
\end{equation}

In our experiments, the initial attraction $\mathcal{A}^j_i(0)$ is implemented according to the Poisson distribution function $P(k) = \frac{e^{-\tau}\tau^k}{k!}$~\cite{camerer2004cognitive}. We set $N(0)=1$,$\kappa=0$\footnote{$\kappa$ specifies cumulative ($k=1$) or average reinforcement ($k=0$).}, $\lambda=2.39$ and $\tau=1.5$, which are the reported optimal parameters for n-player guessing games~\cite{CAMERER2002137}.

Since EWA operates directly on formal game representations rather than natural language, we integrate it into our framework by defining a dedicated interpretation function
\[
I_{\text{EWA}} : G \rightarrow \mathbb{Z}, \qquad
I_{\text{EWA}}(G) \sim P_i(t{+}1),
\]
which samples a strategy according to the softmax distribution in Eq.~(8). In this case, the umpire $\upsilon$ is responsible for translating natural language descriptions $x \in X$ into their corresponding formal game representations via a mapping $T : X \rightarrow G$. Accordingly, when EWA agents participate in the framework, the subject of their game requests is the formalised game rather than the natural language description.

\subsection{Artificial Reasoners}

In this section, we introduce 3 categories of agentic and conceptual sophistication of the implementation suitably adapted from~\cite{trencsenyi2025approximating}: (1) the LLM used for reasoning, (2) the model used for composing the LLM prompts and (3) the agentic workflow the LLM is wrapped in.

\paragraph{LLM}
We tested two state-of-the-art LLMs with our artificial reasoners~\cite{claude37sonnet} representing different levels of complexity: (1) Claude \textbf{Haiku} 3.5 is a quick and compact model with strong reasoning capabilities and (2) \textbf{Sonnet} 3.7 is the latest, most capable model, suitable for complex problem-solving.

\paragraph{Prompting Model}
We construct our prompts based on the typical separation: context ($C$) and instruction ($M$). Providing LLMs with a character description aids their reasoning and instruction following~\cite{park2023generativeagentsinteractivesimulacra,mao2024alympicsllmagentsmeet,trencsenyi2025approximating}. We define three context configurations that represent different, increasingly more elaborate details for the LLM-driven agent as $C=\{c_{\varnothing},c_{\text{sim}},c_{\text{bio}}\}$:
\begin{itemize}
    \item \textbf{No profile ($\mathbf{c_{\varnothing}}$):} agents are presented with situational descriptions and are instructed to pick a number without further context;
    \item \textbf{Simple profile ($\mathbf{c_{\text{sim}}}$):} agents are instructed to reason according to a role specification, whether the agent is an expert of game theory or a novice without any expertise;
    \item \textbf{Biography ($\mathbf{c_{\text{bio}}}$):} the Umpire initialises each player with an elaborate background supporting the specific role.
\end{itemize}

Furthermore, we implement an additional configuration parameter based on a psychological model of human decision-making inspired by the agentic model of appropriateness (\textbf{MoA}) from~\cite{vezhnevets2023generative}. We instruct our agents to address three questions in their responses: (1) \textit{What kind of situation is this?}, (2) \textit{What kind of person am I?} and (3)\textit{What should a person like me do in a situation like this?} While the decoupled decision-making capability specified in our multi-agent framework already implements subtasking, having the LLM respond via the three questions integrates an additional layer of chain-of-thought prompting aiding reasoning~\cite{wei2023chainofthoughtpromptingelicitsreasoning}. Furthermore, as the three questions are associated with how humans make decisions, and as LLM training data contains a vast amount of samples of human culture, we adopt the hypothesis that guiding the agent through the human-inspired patterns will support the generated reasoning to better approximate human responses. We define MoA as a binary feature encapsulated in the instruction $M=\{m_0,m_1\}$, where $m_1$ denotes the configuration that implements MoA.

\paragraph{Agentic Concept}

First, implementing the decoupled revision and decision mechanism from~\cite{trencsenyi2025approximating}, we define the \textbf{Reasoner} ($R$) agent as follows. The interpretation function $I:X \times C \times M \rightarrow \mathbb{Z}$ is defined as $I(x,c_i,m_i)=\delta(\rho(x,c_i,m_i), c_i, m_i)$, where
\begin{itemize}
    \item $\rho: X \times C \times M \rightarrow \Xi$ is the reasoning function: $\rho(x,c,m)=\xi_i$, where $\xi_i \in \Xi$ is $i$'s  belief about the opponent, a natural language reasoning based on game description $x$, context $c_i$, and instruction model $m_i$ on what the opponent guess $\hat{a}_{-i}$ is expected to be;
    \item $\delta : \Xi \times C \times M \rightarrow A$ is the decision function: $\delta(\xi_i,c_i,m_i)$, selecting $i$'s guess $a_i^*$ based on the belief $\xi_i$, context $c_i$, and instruction model $m_i$.
\end{itemize}

In contrast, the \textbf{Simple} ($S$) agent implements a one-step reasoning process, reflecting the growing association in recent literature between LLMs and agents through a looser definition of agency. Simple agents invoke \(I(x,c_i,m_i) = \delta'(x,c_i,m_i)\), where the decision function \(\delta': X \times C \times M \rightarrow A\) internally applies a (possibly identity) mapping \(f: X \rightarrow X'\), producing a belief-enriched description \(x' \in X' = X \times \Xi\). The resulting player guess \(a_i^*\) may thus rely either on the original game description or on an inferred belief \(\xi_{-i}\) about the opponent's guess.

\section{Experiments and Discussion}

The following sections provide a detailed analysis of the experimental results, structured around multiple levels of behavioural fidelity. We evaluate agents’ ability to replicate human-like behaviour at two levels of granularity: first, at the population level, assessing how closely agent outputs align with aggregated human behaviour; and second, at the subpopulation level, measuring how well agents distinguish and replicate the behavioural differences between distinct human groups (e.g., students vs. experts). Within each level, we analyze performance across complementary dimensions of behavioural alignment: (1) mean response behaviour via $k$-level approximation; (2) distributional similarity, using Wasserstein distance; and (3) frequency of zero guesses. Finally, we extend our evaluation to an out-of-sample setting by modifying key game parameters to assess how well agents generalise beyond likely training data priors. For each pairwise comparison where mentioned, we first test for variance homogeneity using Levene’s test. We apply either a standard independent samples t-test (equal variances) or Welch’s t-test (unequal variances). If the data is severely skewed (|skewness| > 2), we use the non-parametric Mann–Whitney U test instead.

\subsection{Benchmark Model Performance}

\begin{figure}[ht]
\centering
\includegraphics[width=\linewidth]{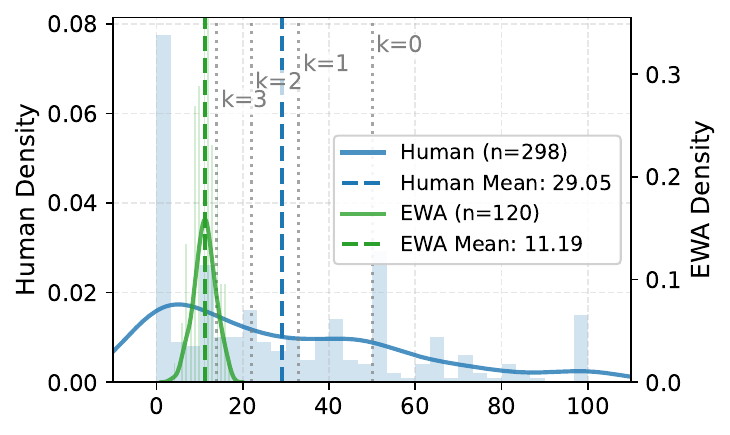}
\caption{Distribution of EWA over joint human results and k-levels.}
\label{fig:ewa}
\end{figure}

As an established game-theoretic model, we first compare the self-tuning EWA agent with the overall human guess, as shown in Figure~\ref{fig:ewa}. The EWA model outperforms human results, averaging around $k=3$. Descriptive statistics confirm this observation: human guesses have a mean of $29.05$ and a standard deviation of $28.28$, while EWA agent guesses average $11.19$ with a standard deviation of $2.50$. Levene's test indicates heterogeneous variances, and a Welch’s $t$-test confirms a statistically significant difference between the groups ($t = 10.80$, $p < 0.0001$). The Wasserstein distance between the two distributions is substantial ($W = 22.34$). The EWA agent produces no zero guesses, indicating that while the EWA model achieves highly strategic behaviour, it systematically diverges from human reasoning patterns, failing to replicate key empirical properties of human play.

\subsection{Artifical Reasoners}

In the following sections, we test each LLM-based agent configuration across Haiku and Sonnet models, agent types -- $S$ or $R$ -- and variations of $C$ and $M$. We refer to specific agent configurations such that, e.g. $S_{c_\varnothing,m_0}$ denotes the Simple agent with no profile, and no MoA or $R_{c_{bio},m_1}$ denotes the Reasoner agent with the Bio context using MoA. Given that EWA cannot process contextual information, we consider it implements $c_\varnothing,m_0$. The two main test cases are specified by testing the (1) overall human-likedness of agents and (2) the ability to approximate human guesses specific to roles.

\subsubsection{Population-wise Performance}

We first analyse performance at the population level, pooling data across subgroups to represent the overall population of participants. This evaluation captures how well agents approximate the aggregate patterns of human reasoning.

\paragraph{Reasoning Sophistication}

We convert numerical guesses to $k$-levels, allowing us to measure performance in terms of reasoning depth. Following~\cite{camerer2004cognitive}, $k=0$ actors are assumed to act randomly with a mean guess of 50, and reasoning depth is derived by solving
\[
\left(\tfrac{2}{3}\right)^k \cdot 50 = a_i^* .
\]
We treat architecture as the first dimension of sophistication in increasing order: $\text{EWA} < S < R$. Secondly, the sophistication of context/instruction configurations $(C,M)$ is given by the ordering:
\[
c_\varnothing,m_0 < c_\varnothing,m_1 < c_{\text{sim}},m_0 < c_{\text{sim}},m_1 < c_{\text{bio}},m_0 < c_{\text{bio}},m_1 .
\]
Our initial expectation was that increasing sophistication along both axes would monotonically improve alignment with human behaviour.

\begin{figure}[ht]
\centering
\includegraphics[width=\linewidth]{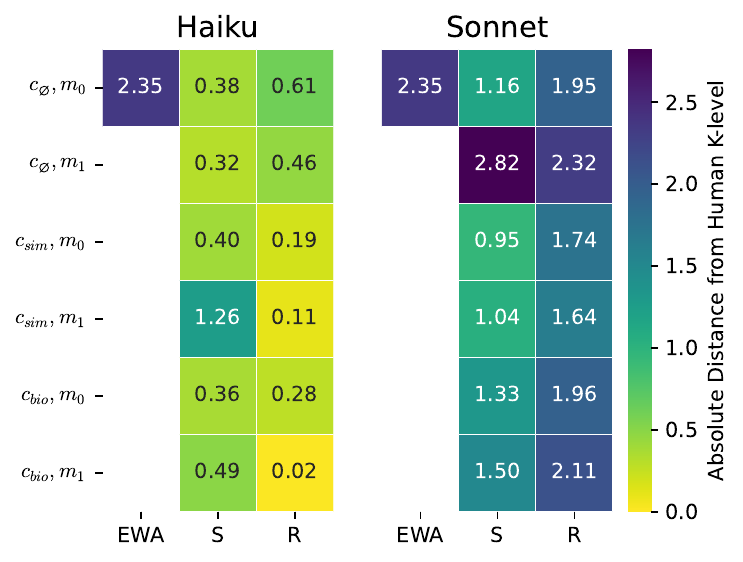}
\caption{Heatmap showing the absolute distance between agents' and humans' mean reasoning sophistication according to k-levels.}
\label{fig:heatmap1}
\end{figure}

Figure~\ref{fig:heatmap1} shows LLM-wise heatmaps of the absolute mean error (AME) $\lvert \bar{k}_{\text{true}} - \bar{k}_{\text{pred}} \rvert$. At the extremes, results align with intuition: EWA is the most distant, while Haiku, with the most elaborate configuration, matches humans almost exactly (AME $=0.02$). However, intermediate cases deviate from this trend: Sonnet, in particular, often performs worse with more sophisticated configurations, and overall diverges more from human reasoners than Haiku.

\paragraph{Factor Analysis}

We transform our categorical features into ordinal variables as follows: agent $\in \{\text{EWA}=0, S=1, R=2\}$, $c \in \{c_{\varnothing}=0,c_{sim}=1,c_{bio}=2\}$, $m \in \{m_0=0 ,m_1=1\}$, and model $\in \{\text{Haiku}=0 , \text{Sonnet}=1\}$. Then, the overall sophistication score is computed as the sum of their scores. Table~\ref{tab:sophistication_effects} illustrates the statistical impact of individual factors on our LLM-based agents, as measured by linear regression models applied to the k-level AME. While the type of agent does not significantly affect the results, $C$ and $M$ show a significant impact. Interestingly, the direction of this effect does not align with our initial expectations. Additionally, a Spearman correlation test reveals a weak and statistically insignificant negative correlation between the sophistication score and the deviation from human k-level performance.

\begin{figure*}[ht]
\centering
\includegraphics[width=\linewidth]{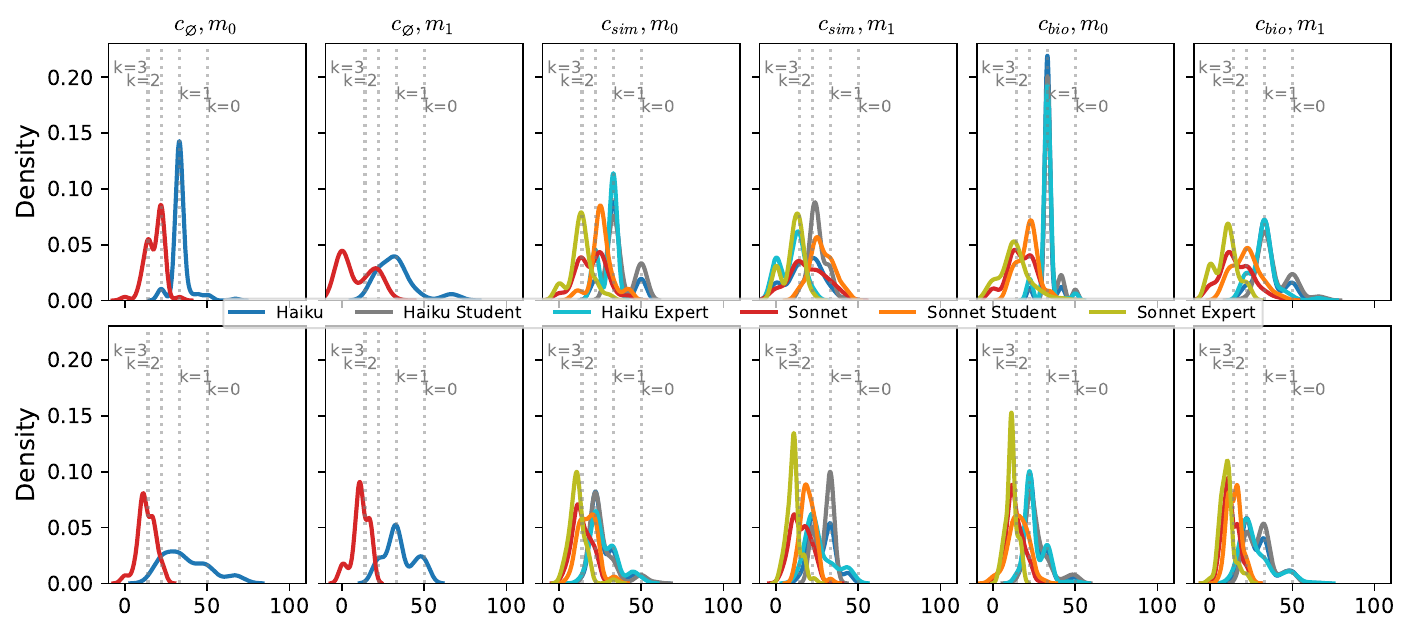}
\caption{KDEs of $S$ (top) and $R$ (bottom) agents with $C$,$M$ variations. Red and blue correspond to overall results, and lighter/darker shades correspond to specific cohorts.}
\label{fig:no_profile}
\end{figure*}

\paragraph{Distributional Similarity}
Since $k$-levels summarise only mean behaviour, we also assess complete guess distributions using kernel density estimates (Figure~\ref{fig:no_profile}) and Wasserstein distance. Tables~\ref{tab:wass_1} and~\ref{tab:sophistication_effects} report the distributional distances and the statistical effects of individual factors. Haiku again generally outperforms Sonnet, and Reasoner ($R$) agents achieve the closest match. Nevertheless, the most human-like outcome is produced by the least sophisticated configuration ($c_\varnothing,m_0$), highlighting that greater sophistication does not guarantee better alignment. Another difference between the LLMs is that MoA consistently benefits Haiku but has a negligible effect on Sonnet.

\begin{table}[ht]
\caption{Statistical effects of sophistication factors and overall sophistication score on $k$-level error, Wasserstein distance, and deviation from human zero-guessing rate.}
\centering
\setlength{\tabcolsep}{4pt}
\begin{tabular}{lrrrrrr}
\toprule
\textbf{Factor} & \multicolumn{2}{c}{$k$-Level Error} & \multicolumn{2}{c}{Wass. Distance} & \multicolumn{2}{c}{0-Rate Error} \\
\cmidrule(lr){2-3} \cmidrule(lr){4-5} \cmidrule(lr){6-7}
                & \textbf{Coef.} & \textbf{p-val} & \textbf{Coef.} & \textbf{p-val} & \textbf{Coef.} & \textbf{p-val} \\
\midrule
Agent Type           & 0.073   & 0.8822  & -0.452  & 0.6484  & -2.609 & 0.1734 \\
LLM Model            & 2.524   & 0.0000  &  0.416  & 0.6748  & -2.609 & 0.1734 \\
$C$    & -0.564  & 0.0186  & -1.543  & 0.2012  & -2.586 & 0.2852 \\
$M$    & 1.272   & 0.0000  & -1.033  & 0.2926  & -1.442 & 0.4585 \\
\midrule
Spearman $\rho$ & -0.025 & 0.1901 & -0.055 & 0.7987 & -0.042 & 0.8447 \\
\bottomrule
\end{tabular}
\label{tab:sophistication_effects}
\end{table}

\begin{table}[ht]
\caption{Wasserstein distances between agents and humans for each mind type across LLMs. Bold values indicate the overall best match per mind (smallest distance), italics mark the best match per LLM, and asterisks highlight the best score in the table.}
\centering
\begin{tabular}{l||c|cc|cc}
\toprule
\makecell{\textbf{Config}} & \textbf{EWA} & \multicolumn{2}{c|}{\textbf{Simple}} & \multicolumn{2}{c}{\textbf{Reasoner}} \\
                & & \textbf{Haiku} & \textbf{Sonnet} & \textbf{Haiku} & \textbf{Sonnet} \\
\midrule
\makecell{$c_\varnothing,m_0$} & 22.34 & 22.05 & 18.59 & \textit{\textbf{14.28*}} & 20.38 \\
\makecell{$c_\varnothing,m_1$} & N.A. & 16.69 & 19.79 & 17.51 & 20.88 \\
\makecell{$c_{\text{sim}},m_0$} & N.A. & 19.92 & 15.41 & 17.46 & 19.32 \\
\makecell{$c_{\text{sim}},m_1$} & N.A. & \textit{15.05} & \textbf{\textit{14.38}} & 17.42 & \textit{18.88} \\
\makecell{$c_{\text{bio}},m_0$} & N.A. & 22.99 & 16.57 & 18.37 & 20.33 \\
\makecell{$c_{\text{bio}},m_1$} & N.A. & 19.08 & 16.24 & 16.14 & 20.78 \\
\bottomrule
\end{tabular}
\label{tab:wass_1}
\end{table} 

\paragraph{Equilibrium Play}
Finally, we evaluate agents on their ability to produce the equilibrium strategy: picking $0$. Human players achieve this in $23.15\%$ of cases. EWA, despite producing higher $k$-levels than humans, never selects zero. As Table~\ref{tab:sonnet_zeros} reveals, Sonnet agents generate zero guesses in most settings, with MoA boosting rates substantially, whereas Haiku rarely produces zeros -- the sole exception being $S_{c_{\text{sim}},m_1}$ at $14\%$. Overall, zero-guessing behaviour does not significantly correlate with the sophistication score.

\begin{table}[ht]
\caption{$0$ guesses per 100 agents (Sonnet only). Human benchmark at $23$.}
\centering
\setlength{\tabcolsep}{3pt}
\begin{tabular}{c|c|c|c|c|c|c}
    \toprule
    \textbf{Agent}  & $\mathbf{c_\varnothing,m_0}$ & $\mathbf{c_\varnothing,m_1}$ & $\mathbf{c_{\text{sim}},m_0}$ & $\mathbf{c_{\text{sim}},m_1}$ & $\mathbf{c_{\text{bio}},m_0}$ & $\mathbf{c_{\text{bio}},m_1}$  \\
    \midrule
    $S$ & $2$ & $52$ & $5$ & $10$ & $9$ & $11$ \\
    $R$ & $4$ & $8$ & $0$ & $0$ & $1$ & $1$ \\
    \bottomrule
\end{tabular}
\label{tab:sonnet_zeros}
\end{table}

\subsubsection{Subpopulation-wise Performance}
Simulations that require an individual-based approach, where specific subpopulations can be represented explicitly, must implement role specification. The following analysis concerns the configurations supporting profile assignment and investigates how well artificial agents can replicate the intergroup differences, role-specific reasoning levels, and guess distributions. As shown in Figure~\ref{fig:humans}, human experts exhibit a substantially lower mean guess and greater density over lower values than students. These results provide the benchmark for our subpopulation-level evaluation.

\paragraph{Reasoning Sophistication}

\begin{figure}[ht]
\centering
\includegraphics[width=\linewidth]{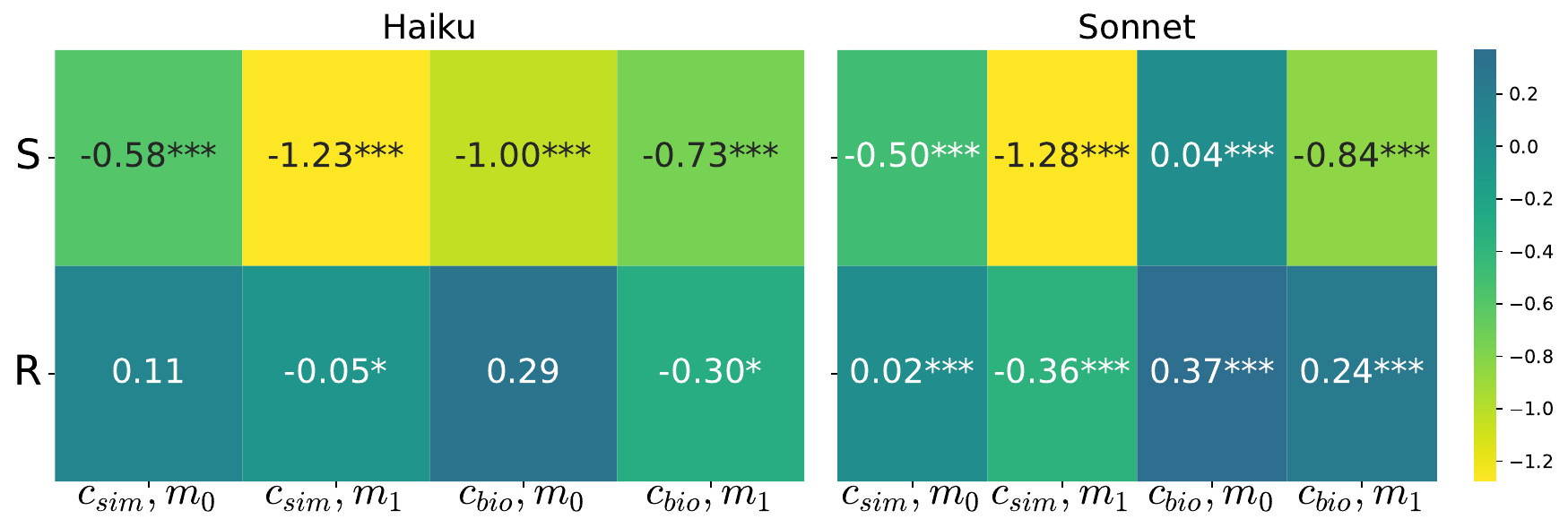}
\caption{Delta error between agents' and humans' mean k-level difference for students and experts with significance levels: * p < 0.05, ** p < 0.01, *** p < 0.001.}
\label{fig:heatmap_profiles}
\end{figure}

First, we calculate the k-level means for students and experts separately and define $\Delta = \left| \bar{k}_{\text{true}}^{\text{student}} - \bar{k}_{\text{pred}}^{\text{student}} \right| - \left| \bar{k}_{\text{true}}^{\text{expert}} - \bar{k}_{\text{pred}}^{\text{expert}} \right|$ to measure whether an agent approximates both groups equally well, with values close to zero indicating better replication of intergroup differences. Figure~\ref{fig:heatmap_profiles} entails the $\Delta$ heatmaps for Haiku and Sonnet. The dominantly negative values in the case of $S$ agents across both models indicate that these agents are consistently closer to expert behaviour, likely due to an overestimation of student reasoning levels, which is likely an overfit to game-theoretic optimizations we observed in previous instances. In contrast, $R$ agents show more balanced behaviour, with errors within $\pm 0.4$ $k$-levels across all $C \times M$ settings. Transitioning from $m_0$ to $m_1$ consistently increases optimisation, raising predicted reasoning levels. Statistical tests confirm that divergence is significantly lower for $R$ agents using Haiku, with particularly close Delta matches in $R_{c_{\text{sim}},m_0}$ and $R_{c_{\text{bio}},m_0}$.

\paragraph{Distributional Similarity}
We extend the analysis to distributional alignment, adding profile-specific KDE curves to Figure~\ref{fig:no_profile} and reporting Wasserstein distances in Table~\ref{tab:wass_profile}. For Haiku, the inclusion of MoA correlates with lower distances, whereas Sonnet exhibits inconsistent patterns without clear trends. The relative overoptimization of Sonnet and its poorer instruction following ability leads to a similar resistance to increasing complexity as we have seen in the experiments without profile specification. In three of four cases, Haiku paired with $c_{\text{sim}},m_1$ achieves the closest match. In the case where Haiku did worse, it produced an outlier distribution with unusually low variance and a concentration around $k=1$ and $k=2$, suggesting that the model latched onto canonical game-theoretic reasoning patterns. While a manual inspection of the produced reasoning samples supports this hypothesis, a deeper semantic analysis of reasoning traces lies outside the scope of this paper.

\begin{table}[ht]
\caption{Wasserstein distances between agent and human guess distributions for each agent type across LLMs. Bold values indicate the overall best match per agent (smallest distance) for the given cohort; italics mark the best match per LLM, and * highlights the best score in the table.}
\centering
\setlength{\tabcolsep}{3pt}
\begin{tabular}{l||cc|cc||cc|cc}
\toprule
\multirow{3}{*}{\textbf{Config}} & \multicolumn{4}{c||}{\textbf{Student}} & \multicolumn{4}{c}{\textbf{Expert}} \\
\cmidrule{2-9}
& \multicolumn{2}{c|}{\textbf{Simple}} & \multicolumn{2}{c||}{\textbf{Reasoner}} & \multicolumn{2}{c|}{\textbf{Simple}} & \multicolumn{2}{c}{\textbf{Reasoner}} \\
\cmidrule{2-9}
& \textbf{H} & \textbf{S} & \textbf{H} & \textbf{S} & \textbf{H} & \textbf{S} & \textbf{H} & \textbf{S} \\
\midrule
\makecell{$c_{\text{sim}},m_0$} & 16.82 & 18.6 & 18.47 & 20.70 & 23.65 & 16.37 & 18.60 & \textit{17.40} \\
\midrule
\makecell{$c_{\text{sim}},m_1$} & 18.74 & 16.98 & 18.16 & \textit{21.04} & \textit{14.68} & 15.84 & 17.33 & 18.02 \\
\midrule
\makecell{$c_{\text{bio}},m_0$} & 20.39 & 18.63 & 18.01 & 21.50 & 25.36 & \textbf{\textit{14.31*}} & 18.82 & 18.28 \\
\midrule
\makecell{$c_{\text{bio}},m_1$} & \textbf{\textit{16.10}} & \textit{16.93} & \textbf{\textit{15.69*}} & 23.14 & 22.69 & 15.59 & \textbf{\textit{17.22}} & 17.54 \\
\bottomrule
\end{tabular}
\label{tab:wass_profile}
\end{table}

\paragraph{Equilibrium Play}

We next investigate how the zero guesses are spread over the subpopulations. As Table~\ref{tab:zeros_profiles}, although overall rates remain sparse, experts predictably guess zero more frequently than students, consistent with the human data, where experts were three times more likely to play the dominant strategy.

\begin{table}[ht]
\caption{Zero guesses per specified profile. Bold values are the cohort-wise best match ($11.9$ and $37.69$) per agent type, italic values denote the best match per LLM and * marks the best score in the table.}
\centering
\setlength{\tabcolsep}{3pt}
\begin{tabular}{l||cc|cc||cc|cc}
\toprule
\multirow{3}{*}{\textbf{Config}} & \multicolumn{4}{c||}{\textbf{Student}} & \multicolumn{4}{c}{\textbf{Expert}} \\
\cmidrule{2-9}
& \multicolumn{2}{c|}{\textbf{Simple}} & \multicolumn{2}{c||}{\textbf{Reasoner}} & \multicolumn{2}{c|}{\textbf{Simple}} & \multicolumn{2}{c}{\textbf{Reasoner}} \\
\cmidrule{2-9}
& \textbf{H} & \textbf{S} & \textbf{H} & \textbf{S} & \textbf{H} & \textbf{S} & \textbf{H} & \textbf{S} \\
\midrule
\makecell{$c_{\text{sim}},m_0$} & 0 & 0 & 0 & 0 & 0 & 10& 0 & 0\\
\midrule
\makecell{$c_{\text{sim}},m_1$} & 0 & 0 & 0 & 0 & 28& 20& 0 & 0 \\
\midrule
\makecell{$c_{\text{bio}},m_0$} & 0 & 0 & 0 & 2& 0 & 18& 0 & 0 \\
\midrule
\makecell{$c_{\text{bio}},m_1$} & 0 & 0 & 0 & 0 & 0 & 22& 0& 2\\
\bottomrule
\end{tabular}
\label{tab:zeros_profiles}
\end{table}

\paragraph{Factor Analysis}

\begin{table}[ht]
\caption{Cohort-specific regression coefficients and Spearman correlations for sophistication influence across three behavioural alignment metrics. Stars indicate statistical significance: * $p < .05$, ** $p < .01$, *** $p < .001$.}
\centering
\setlength{\tabcolsep}{2pt}
\begin{tabular}{l|ccc||ccc}
\toprule
\textbf{Factor} & \multicolumn{3}{c||}{\textbf{Student}} & \multicolumn{3}{c}{\textbf{Expert}} \\
                & $k$-Err. & Was. Dist. & 0 Err. & $k$-Err. & Was. Dist. & 0 Err. \\
\midrule
Agent           & $-3.80^{***}$ & $-1.69$      & $0.25$     & $-6.90^{***}$ & $0.66$       & $-12.00^{**}$ \\
LLM   & $0.97^{***}$  & $1.89$       & $-0.25$    & $2.46^{***}$  & $-3.13^{*}$  & $-5.50$       \\
$C$        & $-0.10$       & $-0.11$      & $0.25$     & $0.73^{*}$    & $-0.99$      & $-2.00$       \\
$M$    & $-0.03$       & $-0.79$      & $0.25$     & $1.51^{***}$  & $-1.73$      & $-5.50$       \\
\midrule
$\rho$      & $0.19^{***}$  & $0.01$       & $-0.18$    & $-0.08^{*}$   & $0.03$       & $0.21$        \\
\bottomrule
\end{tabular}
\label{tab:regression_combined_with_rho}
\end{table}

Table~\ref{tab:regression_combined_with_rho} summarises regression coefficients and Spearman correlations for sophistication factors across three behavioural metrics. The results support earlier observations. Despite the sparsity of zero-guessing behaviour, we find that $R$ agents align significantly better with expert zero-guess rates. For distributional similarity (Wasserstein distance), most factors show no significant effect, with the notable exception of LLM choice, which strongly affects experts and shows a trend for students ($p=0.07$). Agent type and LLM model have a strong and consistent influence on $k$-level approximation for both cohorts: more complex reasoning agents using the smaller LLM (Haiku) achieve lower error. In contrast, contextual ($C$) and instruction complexity ($M$) have only minor and non-significant effects for students, while for experts, increased complexity correlates with higher $k$-level error.
Finally, the Spearman $\rho$ correlation analysis reveals a significant but modest correlation between overall sophistication and $k$-level error: positive for students, suggesting over-complexity harms alignment, and negative for experts, indicating modest improvements. No monotonic trends emerge for distributional similarity or equilibrium play.

\subsection{Out-of-Sample Validation}

Most traditional guessing game experiments involve $n$ players selecting numbers between $0$ and $100$~\cite{NAGEL2008391}. In contrast, our 2-player variant may be underrepresented or entirely absent from LLM training. We treat our setup as an out-of-sample case and further examine generalization by shifting key parameters to diverge from training data priors.

\paragraph{Shifted Range}

Humans are expected to understand the game regardless of varying parameters~\cite{costagomes2006guessinggames}. We test whether artificial agents can do the same by shifting the range of valid guesses to $[100, 200]$, following~\cite{ho-camerer-weigelt1998guessinggames-100-to-200}. This allows us to assess whether our artificial reasoners are misled by their training data where this problem is underrepresented or can identify the distinct game parameters and adapt their strategy to novel numerical boundaries.

\begin{figure}[ht]
\centering
\includegraphics[width=\linewidth]{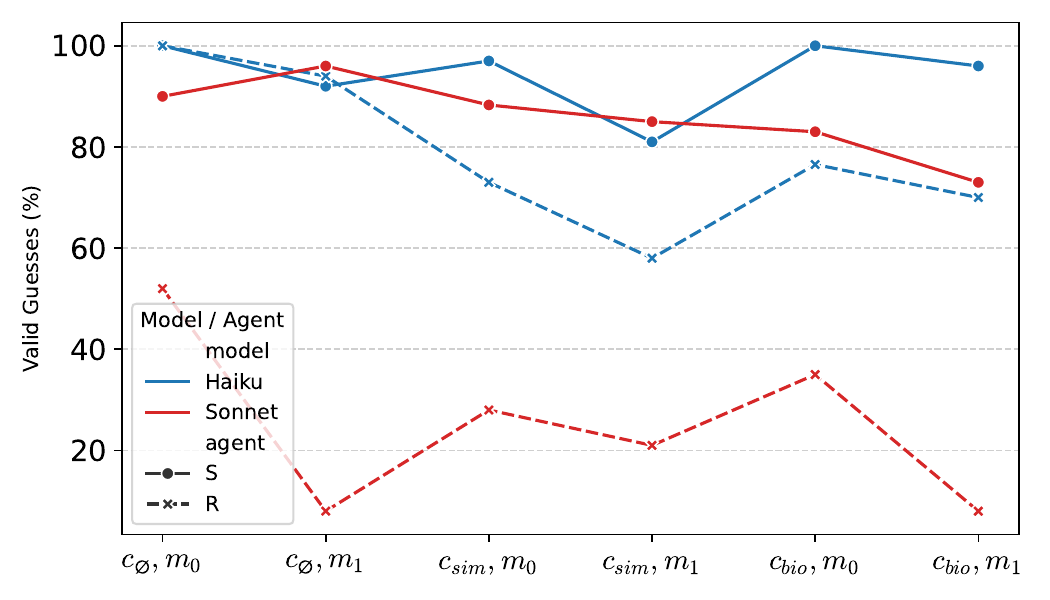}
\caption{Trend of ratios of valid guesses across increasingly complex configs and agent models.}
\label{fig:trend_scaled}
\end{figure}

As with the standard range, the smaller model, Haiku, outperforms Sonnet in its ability to generalise for the modified parameters and provide valid responses -- as depicted by Figure~\ref{fig:trend_scaled}. We can also observe a trend showing that simpler implementations can generalise better. In addition to the performance gap between Haiku and Sonnet, the transition from $S$ to $R$ also decreases the scores. The inclusion of MoA results in a drop in accuracy in each case. Except for the Simple agent using Haiku, the increasing complexity in $C$ shows a negative correlation with guess validity.

Assuming human understanding of the game is invariant to the number range, we normalise shifted-range guesses by subtracting 100 and compare them to standard-range distributions. We then compute Wasserstein distances and t-statistics between original and scaled conditions to identify robustness across agent types and configurations. The results in Table~\ref{tab:wasserstein_tstats} reinforce earlier findings: Haiku does better than Sonnet, and $R$ agents are more dissimilar. While increasing the $C$ complexity does not lead to significant and consistent changes, $m_1$ increases the distance score in each case over $m_0$.

\begin{table}[ht]
\caption{Wasserstein distances (W) and t-statistics (t) comparing KDEs for original vs. scaled guesses. Asterisks mark statistically significant t-values ($p < 0.05$). The closest significant matches per agent type are in bold.}
\centering
\setlength{\tabcolsep}{2pt}
\renewcommand{\arraystretch}{1.1}
\begin{tabular}{l||cc|cc|cc|cc}
\toprule
\multirow{3}{*}{\textbf{Config}} &
\multicolumn{4}{c|}{\textbf{Simple}} &
\multicolumn{4}{c}{\textbf{Reasoner}} \\
\cmidrule{2-9}
& \multicolumn{2}{c|}{\textbf{Haiku}} & \multicolumn{2}{c|}{\textbf{Sonnet}} &
  \multicolumn{2}{c|}{\textbf{Haiku}} & \multicolumn{2}{c}{\textbf{Sonnet}} \\
\cmidrule{2-9}
& \textbf{W} & \textbf{t} & \textbf{W} & \textbf{t} & \textbf{W} & \textbf{t} & \textbf{W} & \textbf{t} \\
\midrule
\makecell{$c_\varnothing,m_0$} & 2.70 & 0.02 & 15.99 & 12.34* & \textbf{7.69} & \textbf{2.77*} & 13.20 & 18.36* \\
\makecell{$c_\varnothing,m_1$} & 9.06 & \textbf{2.10*} & 9.26 & 6.39* & 13.62 & 5.87* & 27.84 & -0.55 \\
\makecell{$c_{\text{sim}},m_0$} & 7.21 & 4.70* & 12.19 & 8.61* & 7.60 & 1.54 & 13.63 & 14.41* \\
\makecell{$c_{\text{sim}},m_1$} & 8.81 & -4.06* & 12.64 & 8.33* & 14.76 & 7.00* & 14.78 & 5.12* \\
\makecell{$c_{\text{bio}},m_0$} & \textbf{5.48} & 5.46* & 14.05 & 12.08* & 8.39 & 1.04 & 12.29 & 15.03* \\
\makecell{$c_{\text{bio}},m_1$} & 7.34 & 4.38* & 13.79 & 11.80* & 13.38 & 6.31* & 12.36 & 26.41* \\
\bottomrule
\end{tabular}
\label{tab:wasserstein_tstats}
\end{table}

\subsection{Cost of Complexity}

Given that $R$ agents involve an additional round of interaction with the LLM per player and that the MoA and the increased amount of details in the base configurations cause lengthier requests and potentially longer responses, in this section, we summarise the estimated incurred costs in terms of input and output token counts in Table~\ref{tab:tokens}.

\begin{table}[ht]
\caption{Cost of LLM agent sophistication in terms of tokens.}
\centering
\setlength{\tabcolsep}{4pt}
\begin{tabular}{ll||cccccc}
\toprule
 & & \makecell{$c_\varnothing,m_0$} & \makecell{$c_\varnothing,m_1$} & \makecell{$c_{\text{sim}},m_0$} & \makecell{$c_{\text{sim}},m_1$} & \makecell{$c_{\text{bio}},m_0$} & \makecell{$c_{\text{bio}},m_1$} \\
\midrule
\multirow{2}{*}{S} 
  & In  & 106.3 & 166.0 & 161.8 & 200.9 & 410.5 & 447.9 \\
  & Out & 189.3 & 311.3 & 256.1 & 414.5 & 243.0 & 330.1 \\
\midrule
\multirow{2}{*}{R} 
  & In  & 292.9 & 390.0 & 341.7 & 642.4 & 878.7 & 935.6 \\
  & Out & 503.7 & 632.9 & 530.8 & 666.1 & 547.9 & 649.0 \\
\bottomrule
\end{tabular}
\label{tab:tokens}
\end{table}

\section{Conclusions}
We conducted an in-depth analysis of the relationship between agentic sophistication and human-likeness in strategic reasoning comparing LLM-driven artificial reasoners with human-data from 2-player guessing games. We used a novel multi-agent-based simulation framework with high conceptual integrity to host agents with varying levels of conceptual complexity representing both looser and stronger agentic definitions, integrating state-of-the-art compact and flagship LLMs. We demonstrated that human-inspired cognitive architectures can successfully mimic human strategic reasoning, but the relationship between implementation complexity and human-likeness is non-linear and is objective-dependent. In addition, our out-of-sample validation with the shifted range of guesses revealed that the compact LLM and simpler configurations generalised better. The evaluation of both main objectives and the robustness test provided consistent insights, which suggest that the larger language model is likely more exposed to overfitting and, thus, generalises worse. These results suggest that researchers should identify the correct models, concepts and architectures to match their specific use case requirements. Defaulting to convenient, simple agentic concepts can lead to suboptimal results; similarly, highly conceptualised implementations may sidetrack agents from the target behaviour.

Promising directions for future research that emerge from our work involve expanding the test beyond a guessing game and evaluating the generalizability of agents across different game-theoretic scenarios. In addition, an in-depth semantic analysis of agents' reasoning processes and a comparison of human and agent natural language reasoning traces and emergent patterns will provide valuable insights into the observed effects of increased complexity and will help researchers design objective-specific artificial reasoning agents.

%%%%%%%%%%%%%%%%%%%%%%%%%%%%%%%%%%%%%%%%%%%%%%%%%%%%%%%%%%%%%%%%%%%%%%%%

% \section{Citations and references}

% Include full bibliographic information for everything you cite, 
% be it a book \citep{pearl2009causality}, a journal article 
% \citep{grosz1996collaborative,rumelhart1986learning,turing1950computing}, 
% a conference paper \citep{kautz1992planning}, or a preprint 
% \citep{perelman2002entropy}. The citations in the previous sentence are 
% known as \emph{parenthetical} citations, while this reference to the 
% work of \citet{turing1950computing} is an \emph{in-text} citation.
% The use of \BibTeX\ is highly recommended. 

%%%%%%%%%%%%%%%%%%%%%%%%%%%%%%%%%%%%%%%%%%%%%%%%%%%%%%%%%%%%%%%%%%%%%%%%

%%% Use this environment to include acknowledgements (optional).
%%% This will be omitted in doubleblind mode.

\begin{ack}
This work was supported by a Leverhulme Trust International Professorship Grant (LIP-2022-001).\\
The authors would like to express their gratidude to David Levine for his constructive comments and fruitful discussions about the game-theoretic material and the EWA model.
\end{ack}

%%%%%%%%%%%%%%%%%%%%%%%%%%%%%%%%%%%%%%%%%%%%%%%%%%%%%%%%%%%%%%%%%%%%%%%%

%%% Use this command to include your bibliography file.

%TODO: clean the references (remove unused + get rid of links or other uglities)
\bibliography{mybibfile}

\end{document}